\documentclass[review]{elsarticle}

\usepackage{lineno,hyperref}
\modulolinenumbers[5]

\journal{Engineering Applications of Artificial Intelligence}

\usepackage{amsmath}
\usepackage{amssymb}
\graphicspath{{figures/}}
\usepackage[usenames, dvipsnames]{color}
\usepackage{multirow}
\usepackage{float}
\usepackage{multirow}
\usepackage{booktabs, caption, makecell}
\usepackage{threeparttable}
\usepackage{url}
\usepackage{amsfonts}
\let\vec\mathbf

\begin{document}
%

\begin{frontmatter}

\title{Attention-Based Sensor Fusion for Human Activity Recognition Using IMU Signals}

\author[a]{Wenjin Tao
	\corref{mycorrespondingauthor}}
\ead{w.tao@mst.edu}

\author[a]{Haodong Chen}
\author[b]{Md Moniruzzaman}

\author[a]{Ming C. Leu}

\author[c]{Zhaozheng Yin}

\author[d]{Ruwen Qin}

\cortext[mycorrespondingauthor]{Corresponding author}

\address[a]{Department of Mechanical and Aerospace Engineering, Missouri University of Science and Technology, Rolla, MO 65409, USA}
\address[b]{Department of Computer Science, Stony Brook University, Stony Brook, NY 11794, USA}
\address[c]{Department of Biomedical Informatics \& Department of Computer Science, Stony Brook University, Stony Brook, NY 11794, USA}
\address[d]{Department of Civil Engineering, Stony Brook University, Stony Brook, NY 11794, USA}

\begin{abstract}
	Human Activity Recognition (HAR) using wearable devices such as smart watches embedded with Inertial Measurement Unit (IMU) sensors has various applications relevant to our daily life, such as workout tracking and health monitoring. In this paper, we propose a novel attention-based approach to human activity recognition using multiple IMU sensors worn at different body locations. Firstly, a sensor-wise feature extraction module is designed to extract the most discriminative features from individual sensors with Convolutional Neural Networks (CNNs). Secondly, an attention-based fusion mechanism is developed to learn the importance of sensors at different body locations and to generate an attentive feature representation. Finally, an inter-sensor feature extraction module is applied to learn the inter-sensor correlations, which are connected to a classifier to output the predicted classes of activities. The proposed approach is evaluated using five public datasets and it outperforms state-of-the-art methods on a wide variety of activity categories.
\end{abstract}


\end{frontmatter}


\section{Introduction}
\label{intro}

Human Activity Recognition (HAR) aims to automatically recognize various human activities, such as daily life and sport activities, with algorithms using the input of a series of sensor measurements.
It has a wide range of applications, such as human-computer interaction, robot learning, ubiquitous computing, workout tracking,
and health monitoring~\cite{casale4, lara2013survey, shoaib3, luo2018computer}.
Although HAR is not a new emerging topic and has been studied for decades, it is still an active area of research now because of remaining challenges, such as
the high complexity of human activities,
the large variations among different subjects,
and the balance between the algorithm complexity and the energy efficiency.

Various sensors have been used for HAR. Considering the wearability, they can be categorized as ambient sensors and wearable sensors.
Ambient sensors are deployed in the environment to sense the subject in a passive manner. For example, optic cameras can be used to capture RGB images on human subjects; Depth cameras such as a Microsoft Kinect or Lidar (light detection and ranging) sensors can be applied to sense human objects in the 3D space; Infrared cameras can detect the subject in a dark environment; Pressure sensing mats can be used to capture human's standing states; WiFi signals also have been used for HAR~\cite{jiang2018towards}. Ambient sensing can collect a large amount of data without interfering the subject's activity.

Nevertheless, ambient sensors require complex setups and their performance can be affected dramatically by occlusion issues, which are the main challenges in implementing ambient sensing. Also, it becomes more difficult when capturing a subject's outdoor activities. 
To compensate for these limitations, wearable sensing can be applied. 
Wearable sensor based activity recognition has captured growing attention nowadays because of the pervasiveness of mobile devices (e.g., smart phones and smart watches), which are embedded with various sensors such as IMU (Inertial Measurement Unit) sensors, heart rate sensors, and ECG  (Electrocardiogram)  sensors. 
%
IMU sensors are the most used for HAR as the sensor directly measure the movements of human body. Usually, an IMU has multiple sensors in different modalities, such as an accelerometer, a gyroscope, and a magnetometer, to measure the acceleration, angular rate, and magnetic field, respectively.

In this paper, we focus on accurately recognizing human's physical activities with multiple IMU sensors considering that IMU signals from different locations could augment the perception of human activities.

\begin{figure}
	\centering
	\includegraphics[width=\columnwidth]{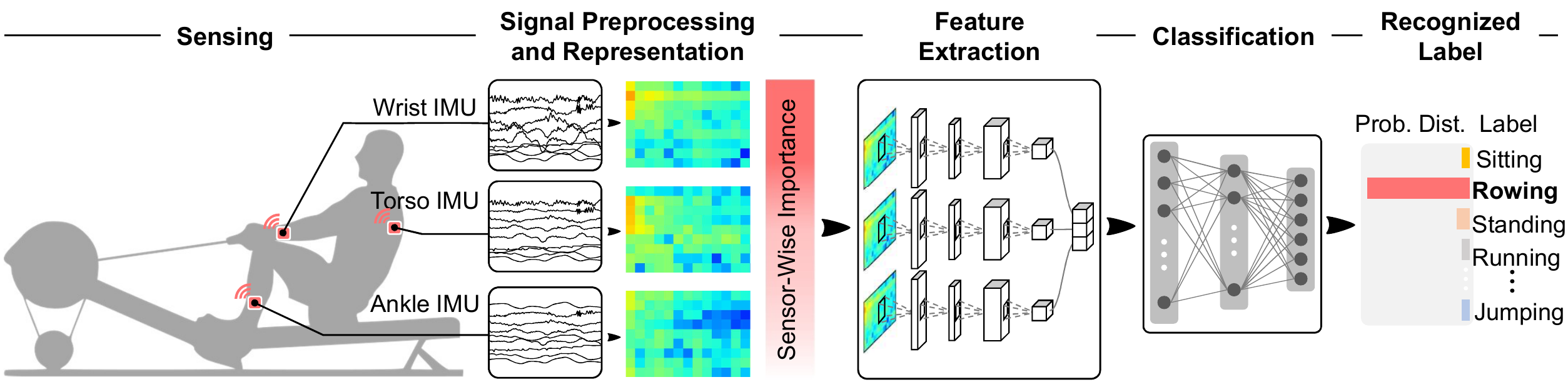}
	\caption{Overview of the human activity recognition pipeline using IMU signals.}
	\label{fig:idea}
\end{figure}

The pipeline of human activity recognition is illustrated in Figure~\ref{fig:idea}. 
IMU sensors are worn at different body locations to sense the activity, from which a series of signals are captured and preprocessed to have formatted representations. After that, a feature extraction process is implemented to extract high-level features. Then, the extracted features are fed into a classifier %
to generate a probability distribution of activity classes. Finally, the activity label can be inferred.

\subsection{Related Work}
\label{related}

The critical factor attributed to the success of IMU-based activity recognition is to seek an effective representation of the time-series IMU signals. The most widely used approaches fall into two categories: handcrafted feature design and automatic feature learning.

\textbf{Hand-Crafted Feature Design.} 
It is intuitive to manually pick statistical attributes (e.g., means) or quantity distributions (e.g., magnitude histograms) from IMU signals~\cite{hosein2016wearable}. For example, Anguita et al. \cite{anguita1} designed as many as 341 features from 3-axis IMU signals while Hammerla et al. \cite{hammerla2013preserving} preserved the statistical characteristics of IMU data using their empirical cumulative distributions. Xu el al. \cite{xu2016learning} proposed a multi-level feature learning framework which consists of the signal-based, components-based and sematic-based information for activity recognition. 
However, handcrafted feature design is mostly driven by the domain knowledge, prior experience and experimental validation, thus it is possible to neglect some useful features in this manner. In addition, a pre-defined feature extraction mechanism trained on a specific scenario might not work well on other scenarios with different sets of activities to be recognized. That is, those hand-crafted features in the literature might not be transferrable to new application domains, which further makes the feature design time-consuming and labor-costly.

\textbf{Automatic Feature Learning.} 
The drawbacks of handcrafted features motivate researchers to explore automatic feature learning \cite{jiang2015human}\cite{ijjina2014one}. Deep Convolutional Neural Network (DCNN), as one of the most effective deep learning models, attracts attentions in the mobile sensing domain considering it has achieved the superior performance in other research fields such as computer vision \cite{krizhevsky2012imagenet} and speech recognition \cite{mohamed2010investigation}. To improve the accuracy of sensor-based activity recognition, Zeng et al. \cite{zeng2014convolutional} designed a tri-thread DCNN architecture with the three inputs corresponding to the tri-axis accelerometry data, thus the inputs are one-dimensional time-series signals. To enhance the ability for feature learning, Duffner et al. \cite{duffner20143d} and Ha et al. \cite{ha2016convolutional} took as input the two-dimensional matrix obtained by stacking IMU signals. In order for further accuracy improvement, Ravi et al.~\cite{ravi2016deep} combined features learned from the deep model with complementary information from a set of hand-crafted features. In addition, Lane et al.~\cite{lane2015can} looked into this problem in a practical way and showed the application of deep learning to mobile sensing domain is hardware-efficient and can scale up to a large number of inference classes.

In short, the input to the deep network and the architecture of the deep model itself are two key factors to the success of automatic feature learning. The input is of great significance because a good representation of the IMU signals can make it easier for automatic learning. In the previous work, IMU signals are directly fed into the DCNN architecture and this simple and raw input may not be a good representation of IMU signals because each value of the raw time-series signals is less meaningful if we do not consider the statisctic property of the whole signals.

In terms of the design of deep architecture, the aforementioned simple input restricts the depth of the deep model, limiting the capability to find discriminative features. For instance, the input in \cite{yang2015deep} is a small $3\times30$ matrix and there are only two convolutional layers in the architecture. Additionally, the tri-axis accelerometry signals are convolved with one-dimensional kernels in the deep model independently, thus the correlation among different signals is not taken into enough consideration.

\textbf{Self-Attention Mechanisms.} 
Just like humans can allocate different amount of attention to different aspects when performing a complex task,  
self-attention mechanisms can model attentions for deep neural networks and have been widely applied in many deep learning tasks~\cite{chorowski2015attention}. The self-attention mechanism is proposed in~\cite{vaswani2017attention} for machine translation tasks, in order to distribute different attention over words in a sentence. From then on, attention mechanisms have been increasingly popular in natural language processing (NLP) and computer vision fields, where multiple sources with different importance are involved. For example, Chen et al.~\cite{chen2017sca} uses spatial and channel-wise attention for image captioning, and He et al.~\cite{he2018stnet} applies attention in both the spatial and temporal domains for HAR from videos.

\subsection{Our Proposal}
A single IMU sensor\footnote{An inertial measurement unit (IMU) can include multiple sensors, such as accelerometers, gyroscopes and magnetometers, here we treat an IMU as an integrated \lq sensor\rq\ for simplicity.} collects data only from a specific body location, which may not perform the robust perception under various circumstances, such as when an activity involves multiple body parts or the movements are not captured from the location the IMU is worn. 
Intuitively, multiple IMU sensors have been used to integrate the perception of individual sensors at different body locations for a better understanding of the overall activity.

Traditional methods treat different IMU sensors equally. 
Few attempts have been made to take the importance of different sensors into consideration when developing HAR algorithms, which cannot provide the correct \lq attention\rq\ on IMU sensors for different activities.
In the present research, to achieve a better understanding of how different sensors contribute to the recognition tasks, we focus on the automatic importance learning for fusing sensors at different body locations.

\begin{figure}
	\centering
	\includegraphics[width=\columnwidth]{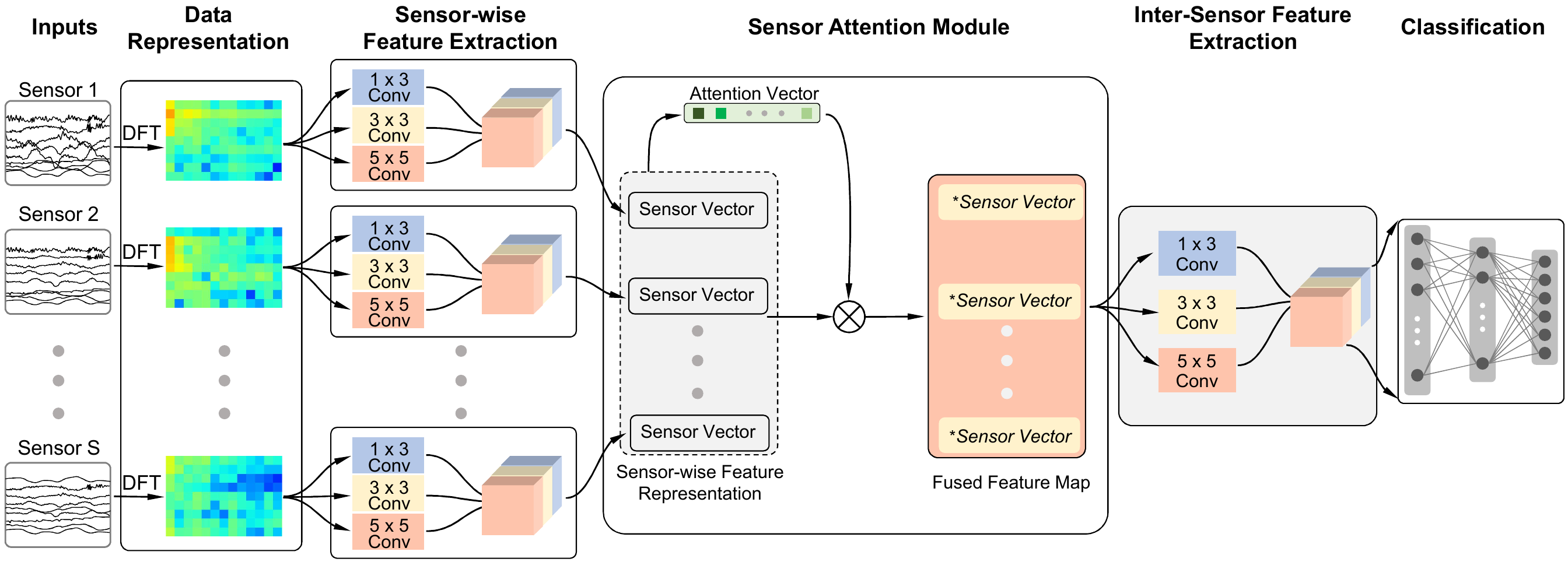}
	\caption{Overview of our attention-based approach for human activity recognition.}
	\label{fig:overview}
\end{figure}

An overview of our approach is illustrated in Figure~\ref{fig:overview}. IMU signals are captured from multiple sensors worn at different body locations. 
Firstly, the signals are preprocessed to generate representations in the frequency domain. 
Secondly, for a sensor at a certain body location, we design a sensor-wise feature extraction module to extract the most discriminative features of signals from each individual sensor.
Thirdly, an attention-based fusion mechanism is developed to learn the importance of sensors at different locations and to generate an attentive feature representation. 
Finally, an inter-sensor feature extraction module is applied to learn the feature relationships among sensors at different locations, which is connected to a classifier to output the predicted classes of activities.
To evaluate our method, five publicly available datasets are chosen which contains a wide variety of activity categories, such as daily activities (sitting, standing, vacuum cleaning, etc.), sports activities (cycling, running, playing basketball etc), and car maintenance activities (opening the hood, etc).

The main contributions of this study are as follows:
\begin{itemize}
	\item Overall, we propose an attention-based approach for human activity recognition using Inertial Measurement Unit (IMU) signals. Multiple IMU sensors are used to perceive the activities and the importance of each individual sensor is automatically learned to achieve an optimal understanding of the human's activities.
	
	\item Regarding to the IMU sensor signal representation, we design a simple yet effective feature transform method to represent the input signals as images in the frequency domain.
	
	\item Regarding to the attention mechanism, we develop a sensor-wise attention module, which enables the network to emphasize features from specific sensors depending on the signals.
	For fusion purpose, multi-kernel convolutional neural networks are applied to extract the most discriminative sensor-wise and inter-sensor features.
	
	\item Regarding to the experimental validation, our approach outperforms other methods on all of the chosen five public datasets.
	
\end{itemize}

The remainder of this paper is organized as follows. 
Section~\ref{method} discusses the details of our proposed approach. Experimental results on five public datasets are described in Section~\ref{experiment}, including comparison with the state-of-the-art methods, and the visualization of the results. Finally, Section~\ref{conclusion} provides the conclusions of this study.

\section{Methods}
\label{method}
In this section, we first present the methods for data preprocessing and representation. Then, each module of our model is explained, including the sensor-wise feature extraction module, sensor attention mechanism, inter-sensor fusion module, and classification module. After that, the training information is detailed.

\subsection{Signal Preprocessing and Representation}
\label{sec:signal_preprocess}
Deep neural networks (DNN) need the input data to be converted as formatted tensors, for example, with a fixed size of $h\times w\times c$ for image inputs where $h$, $w$ and $c$ are the height, width and the number of channels of the image, respectively. Therefore, some preprocessing steps are necessary before the data can be fed into a DNN. 
In this section we give a detailed description of the pipeline for data preprocessing and the methods we use for signal representation.

\begin{figure}[H]
	\centering
	\includegraphics[width=.65\textwidth]{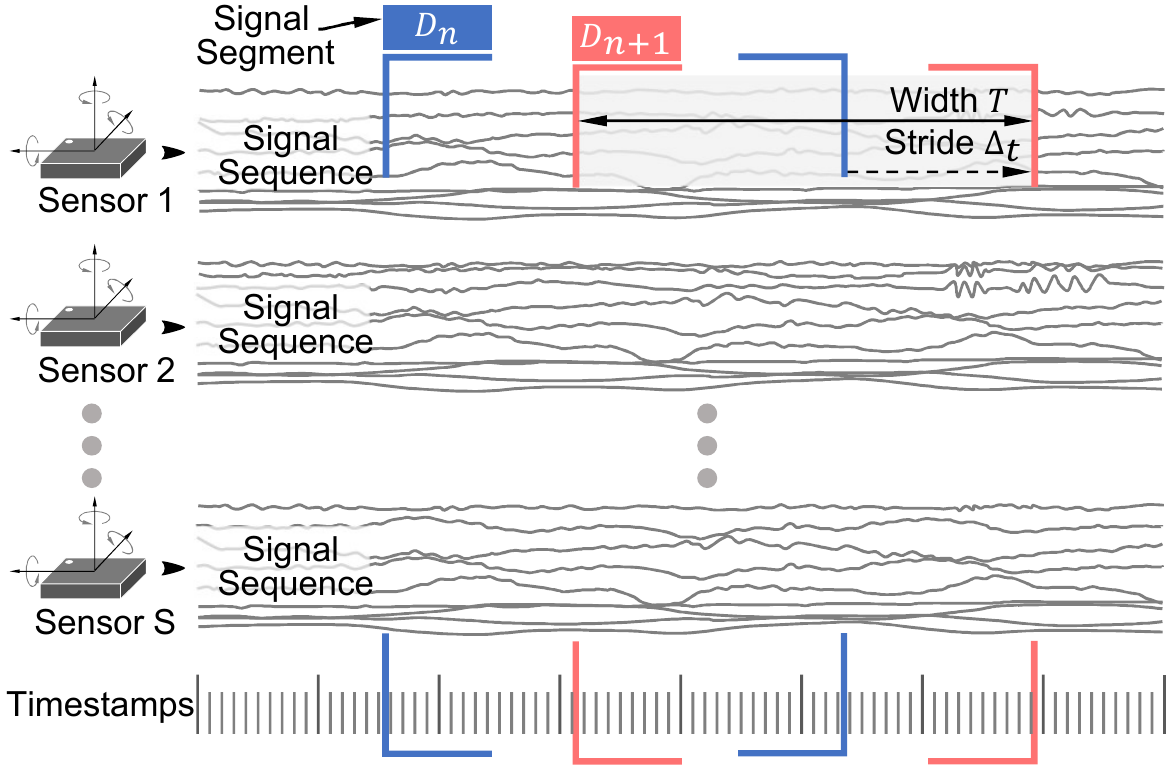}
	\caption{Scheme of the signal sampling method.}
	\label{fig:sample}       
\end{figure}

\textbf{Sampling Procedures.} 
As depicted in Figure~\ref{fig:sample}, the IMU signals from sensors at different body locations are synchronized with the timestamps and denoted as signal sequences. Then, the signal sequences are sampled using a temporal sliding window with the width of $T$ timestamps and $\Delta_t$ stride length between two windows. 

After sampling, we denote our dataset as $\mathbb{D}=\{[D_1, y_1], \cdots, [D_n, y_n], \cdots, [D_N, y_N]\}$ and the $n$th data is represented as
\begin{equation} \label{eq:Dn}
	D_n = [d_n^1, d_n^2, \cdots,d_n^s, \cdots, d_n^S], \quad n\in\{1, \cdots, N\}
\end{equation}
where $S$ is the total number of IMU sensors at different body locations, $d_n^s$ is a sample set of discrete time-series IMU signals 
from the $s$th sensor, 
and $y_n$ is the manually labeled ground truth of the activity class.
More specifically, $d_n^s$ a sequence of discrete-time data over $T$ timestamps, $d_n^s=\{d_{n,1}^s,\cdots,d_{n,t}^s,\cdots,d_{n,T}^s\}$, and each element is elaborated as
\begin{equation}\label{}
	\begin{split}
		d_{n,t}^{s}=[\underbrace{a_{n,t}^x, a_{n,t}^y, a_{n,t}^z}_{a_{n,t}\text{: acceleration}}, \underbrace{g_{n,t}^x, g_{n,t}^y, g_{n,t}^z}_{g_{n,t}\text{: gyro}}, \underbrace{m_{n,t}^x, m_{n,t}^y, m_{n,t}^z}_{m_{n,t}\text{: magnetometer}}, \cdots], \quad
		t\in\{1,\cdots, T\},
	\end{split}
\end{equation}
where ${a}$, ${g}$, and ${m}$ are sensor readings of linear acceleration, angular velocity, and magnetic field, respectively. In some public datasets, derived information such as gravity-removed linear acceleration and orientation in Euler or quaternion form, is also included.

\begin{figure}[H]
	\centering
	\includegraphics[width=.8\textwidth]{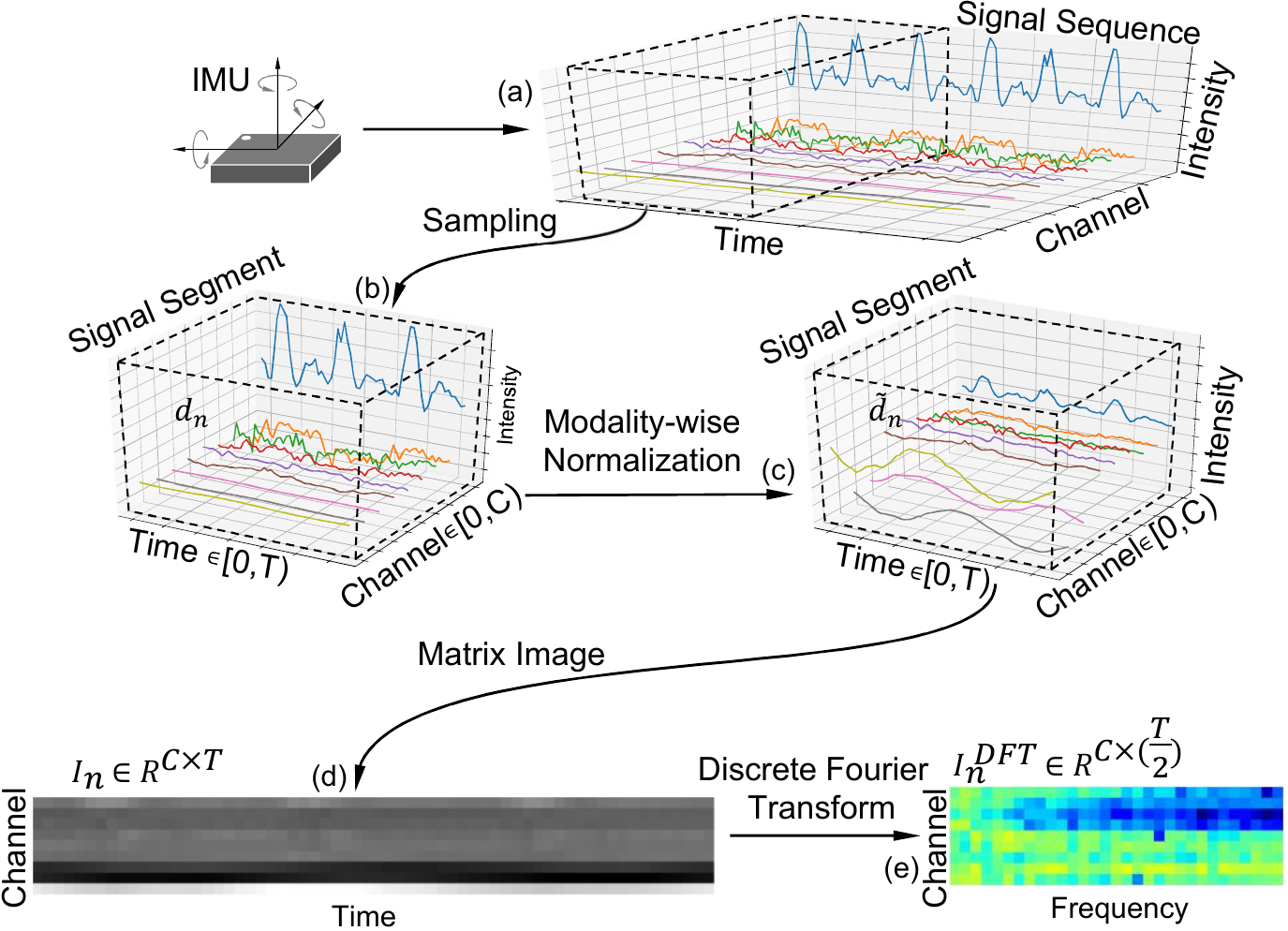}
	\caption{Illustration of the signal representation pipeline for an individual IMU sensor.}
	\label{fig:dft_image}       
\end{figure}

\textbf{Signal Representation.} 
Analyzing signals in the frequency domain is commonly used for signal pattern recognition, because it can extract periodic characteristics which can be more representative than original signals in the time domain. 
In our study, rather than directly modeling the time-series signals with a DNN, frequency transform is applied as follows: 1) As shown in Figure~\ref{fig:dft_image},  a signal segment $d_n$ (Fig.~\ref{fig:dft_image}(b), for simple notation, we drop the superscript $s$ that indicates the $s$th sensor, in the following derivation) is sampled from a signal sequence (Fig.~\ref{fig:dft_image}(a));
2) A modality-wise normalization is applied to $d_n$ to normalize the signal to the range of $[0, 1]$, generating $\tilde{d}_n$ (Fig.~\ref{fig:dft_image}(c)).
3) After normalization, the IMU signal $d_n$ in an IMU segment is represented as an image $I_n^{}$ with the size of $C\times T$ (Fig.~\ref{fig:dft_image}(d)) where $C$ and $T$ denote the numbers of channels and time frames, respectively, resulting in $S$ image representations for all sensors;
4) One-dimensional Discrete Fourier Transform (DFT) along the time dimension 
is applied to $I_n$ to get the representation in the frequency domain for analyzing the frequency characteristics. 
Its logarithmic magnitude is taken to form the image $I_n^{DFT}$. Due to the conjugate symmetry of Discrete Fourier Transforms
\begin{equation}
	\begin{split}
		I_n^{DFT}(k, c)=I_n^{DFT}(-k, c)\, ,\\
	\end{split}
\end{equation}
where $k$ and $c$ represent the two directions (i.e., frequency and signal channel, respectively) of an image $I_n^{DFT}$,  
we can use only a half to represent the DFT image.
In the following, we keep using the notation $I_n^{DFT}$ to represent the one-half of DFT image for simplicity (Fig.~\ref{fig:dft_image}(e)).

Compared with the previous work~\cite{jiang2015human, tao2019multi-modal} for signal representation, our method removes the information redundancy,  
thus reducing the architectural complexity and the number of training parameters for the DNN model.

\begin{figure}[H]
	\centering
	\includegraphics[width=\columnwidth]{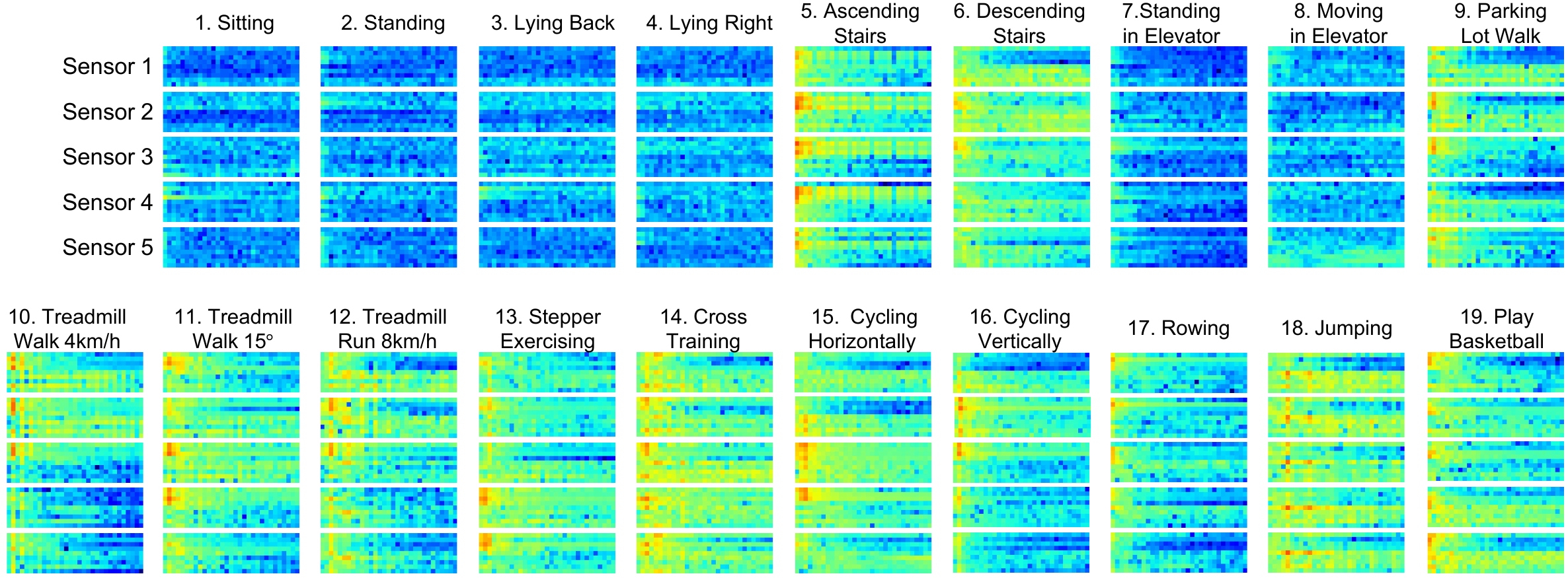}
	\caption{Samples of image representation of different activities from the Daily dataset (5 IMU sensors included).}
	\label{fig:activity_image}
\end{figure}

In total, we have $S$ image representations in the frequency domain for each activity segment. 
For example, five sensors are included in the Daily dataset~\cite{barshan2014recognizing}, i.e., $S=5$.
Figure~\ref{fig:activity_image} shows some examples of image representations in the frequency domain, from one subject on 19 activities, from which we can observe the unique patterns of each activity.

\subsection{Sensor-Wise Feature Extraction Module}
\label{sec:sensor-wise_module}
After the above preprocessing step, we have formatted the input ready for DNN. There are $N$ training data samples $\{X_1, \cdots, X_N\}$, each of which contains $S$ sensor inputs:
\begin{equation}
	X_n = \{I_n^1, \cdots, I_n^s, \cdots, I_n^S\}, \quad n\in[1,N]
\end{equation}

For each of the image inputs $I_n^{s}$, 2D convolutional operation~\cite{Goodfellow-et-al-2016} is applied to extract features layer by layer. The convolutional value using a 2D kernal $K$ at the position $(i,j)$ in the feature map of the $l$th layer 
is computed by
\begin{equation}
	F_{i, j}^l = (F^{l-1}*K)_{i, j}=\sum_{p=0}^{P-1}\sum_{q=0}^{Q-1} F_{i+p,j+p}^{l-1}K_{p,q}
\end{equation}

\noindent where 
$l$ is the layer index, $K_{p,q}$ is the value at the position $(p,q)$ of the kernel, and $P$ and $Q$ are the height and width of the two-dimensional kernel $K$, respectively.

\begin{figure}[H]
	\centering
	\includegraphics[width=.75\textwidth]{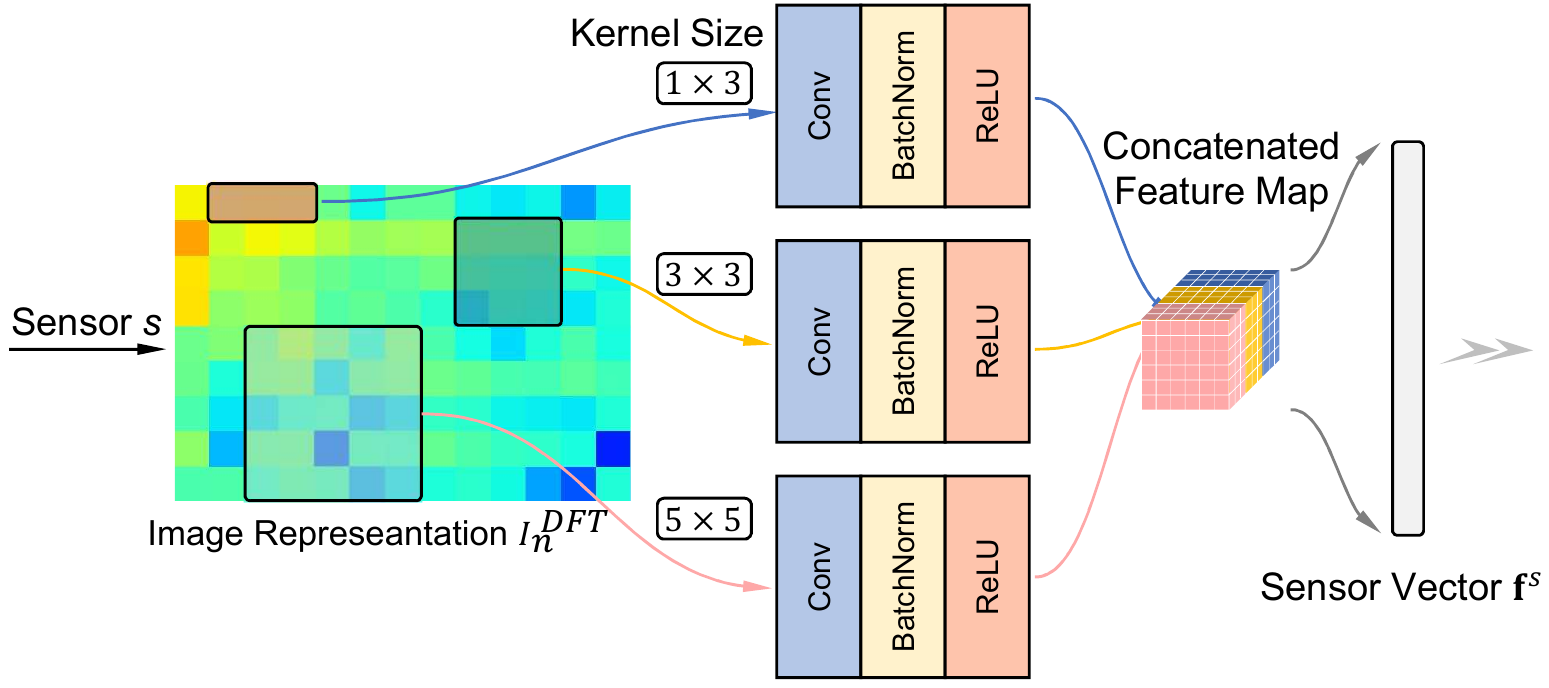}
	\caption{Illustration of the feature extraction module.}
	\label{fig:sensor_block}       
\end{figure}

To learn the hidden correlation patterns among multi-channel signals for each individual sensor, we design an intra-sensor feature extraction module. The motivation is to use multiple convolution kernels with various sizes to detect features across different signal channels. As shown in Figure~\ref{fig:sensor_block}, for the input of the $s$th sensor, $1\times 3$ kernels are used to look at the channel-wise feature, $3\times 3$ kernels are designed to detect the inter-channel features among three channels, and $5\times 5$ kernels are used to discover the inter-channel pattern in a larger perceptive field. In addition, larger size kernels, such as $7\times 7$ and $9\times 9$ can be used to further look into the signals in a larger field. 

After each convolutional layer, a batch normalization layer~\cite{ioffe2015batch} and an activation layer of ReLU (Rectified Linear Unit) are applied.
Then, these extracted feature maps are concatenated to form an information-richer feature set containing features across different signal channels. 
Finally, the extracted feature maps from each sensor is flattened as a vector representation $\vec{f}^s$, which we call a \lq sensor vector\rq\ in the following derivations.

\subsection{Sensor Attention Mechanism}

\begin{figure}[H]
	\centering
	\includegraphics[width=.75\textwidth]{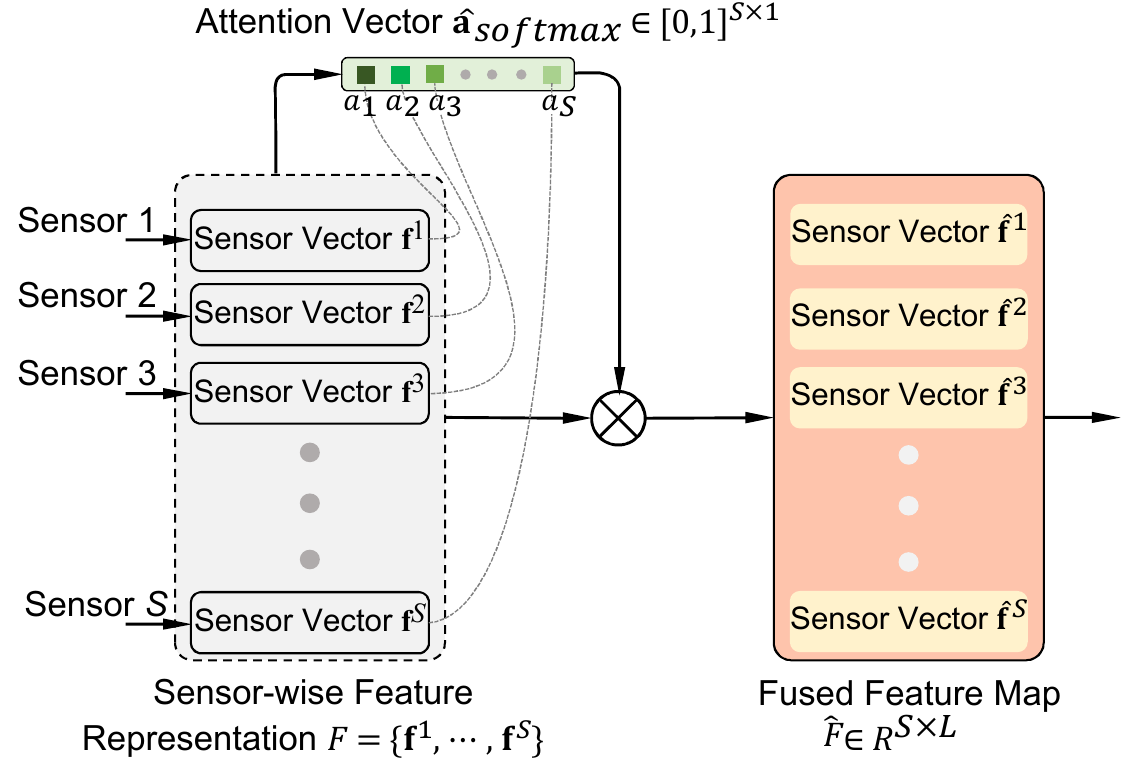}
	\caption{Illustration of the sensor attention mechanism.}
	\label{fig:attention_mechanism}       
\end{figure}

The sensor-wise feature extraction of signals treat every IMU sensor
indiscriminately, but sensors at some body locations may be not or less effective to represent a certain activity and discriminate it from others.
For example, a sensor worn on the ankle may not be able to effectively perceive the \lq rowing\rq\ activity.
Thus, we propose a sensor attention mechanism to learn more attentions on those discriminative sensors in a signal segment. This sensor attention is a trainable layer inside a DNN, which pools the most discriminative features, as shown in Figure~\ref{fig:attention_mechanism}.

Given the sensor-wise feature representation of a signal segment, $F=\{\vec{f}^1, \vec{f}^2,\\\cdots, \vec{f}^s,\cdots, \vec{f}^S\}$, $\vec{f}^s\in R^{L\times 1}$, (where $L$ is the vector dimention and each feature vector is extracted from a sensor within a signal segment), our attention module learns an attention score vector, $\vec{a}$, which indicates the feature importance of different sensors within the signal segment:
\begin{equation}
	\vec{a}=F{w}^{}, \vec{a}\in R^{S\times 1},
\end{equation}
\noindent where $w\in R^{L\times 1}$ is the weight.
Then, the activation vector $\hat{\vec{a}} $ is calculated as
\begin{equation}
	\hat{\vec{a}} = tanh(W \vec{a} + \vec{b}),
\end{equation}
\noindent where $W$ is a weight matrix and $b$ is a bias vector. 

After the activation process, we have a set of attention score $\hat{\vec{a}}=\{a^1, a^2, \cdots, \\a^s, \cdots, a^S\}$. Then, the attention score vector is passed through a \textit{softmax} layer:
\begin{equation}
	\label{eq:attention}
	a^s_{softmax} = \frac{\exp(a^s)}{\sum^S_{s=1}\exp(a^s)}
\end{equation}
to get $\hat{\vec{a}}_{softmax}\in [0,1]^{S\times 1}$.
Then, the attention-applied feature map $\hat{F}$ of the data segment is computed by 
\begin{equation}
	\hat{F} = F\odot \hat{\vec{a}}_{softmax}, \quad \hat{F}\in R^{S\times L}
\end{equation}

\noindent where $\odot$ is the element-wise multiplication operator. Here each sensor (each row in $\hat{F}$) has its corresponding attention-applied feature vector $\hat{\vec{f}}$.

Overall, the proposed sensor attention mechanism fuses inputs from multiple sensors into a single representation by assembling the weighted sensor vectors from individual sensors into a 2D feature map, which enables the network to distribute different amount of attention over different sensors.

\subsection{Inter-Sensor Fusion Module}
\label{sec:inter-sensor}
As shown in Figure~\ref{fig:overview}, after the attention mechanism is applied, each row of the feature map comes from each individual sensor.  The attentive feature map has the size of $S\times L\ (number\ of\ sensors \times dimension\ of\ each\ sensor\ vector)$.  To discover the hidden correlations among different sensors. An inter-sensor fusion module is developed. This module essentially follows the same architecture as presented in Section~\ref{sec:sensor-wise_module}. By using the 2D convolution, the correlation among sensors can be learned.

\subsection{Classification Module}
As shown in Figure~\ref{fig:overview}, a classification module is designed after the inter-sensor fusion module.
First, the feature map obtained from the inter-sensor fusion module are flattened as a feature vector. To solve the classification problem, the vector is further input to a multi-layer neural network. 
The value of the $j$th neuron in the $i$th fully connected layer, denoted as $v_{ij}$, is given by
\begin{equation}
	v_{ij}=g\Bigg(b_{ij}+\sum_{k=0}^{K_{(i-1)}-1}{w_{ijk}v_{(i-1)k}}\Bigg),
\end{equation}
where $b_{ij}$ is the bias term, $k$ indexes the set of neurons in the $(i-1)$th layer connected to the current feature vector, $w_{ijk}$ is the weight value in the $i$th layer connecting the $j$th neuron to the $k$th neuron in the previous layer. 

The last fully connected layer is used to densify the feature vector to the dimensions of $M$, where $
M$ is the number of activity classes. Then this $M$-dimensional score vector $\vec{s} ([s_1, ..., s_m, ..., s_M])$ is transformed to output the predicted probabilities with a softmax function as follows:
\begin{equation}
	P(y_n=m|X_n)=\frac{\exp(s_m)}{\sum^M_{j=1}\exp(s_j)}
\end{equation}
where $P(y_n=m|X_n)$ is the predicted probability of being class $m$ for sample $X_n$.

\subsection{Training}

The process of training a DNN model involves optimization of the network's parameters $\theta$ to minimize the cost function for the training dataset $X$. We select the commonly used regularized cross entropy~\cite{Goodfellow-et-al-2016} as the cost function for the classifier, which is
\begin{equation}
	\mathcal{L}(\theta) = \sum^N_{n=1} \sum^M_{m=1}y_{nm}\log[P(y_n=m | X_n)]+\lambda l_2(\theta)
\end{equation}

\noindent where $y_{nm}$ is 0 if the ground truth label of $X_n$ is the $m$th label, and is 1 otherwise. The $l_2$ regularization term is appended to the loss function for penalizing large weights, and $\lambda$ is its coefficient.

\section{Experiments}
\label{experiment}
In this section, we first describe the selected public datasets and evaluation metrics. Then, we perform evaluation of our proposed approach using these datasets, and compare with the state-of-the-arts. After that, we conduct visualizations for a better understanding of the learned attention. Finally, future research needs are discussed.

\subsection{Datasets}
\begin{figure}[H]
	\centering
	\includegraphics[width=.85\textwidth]{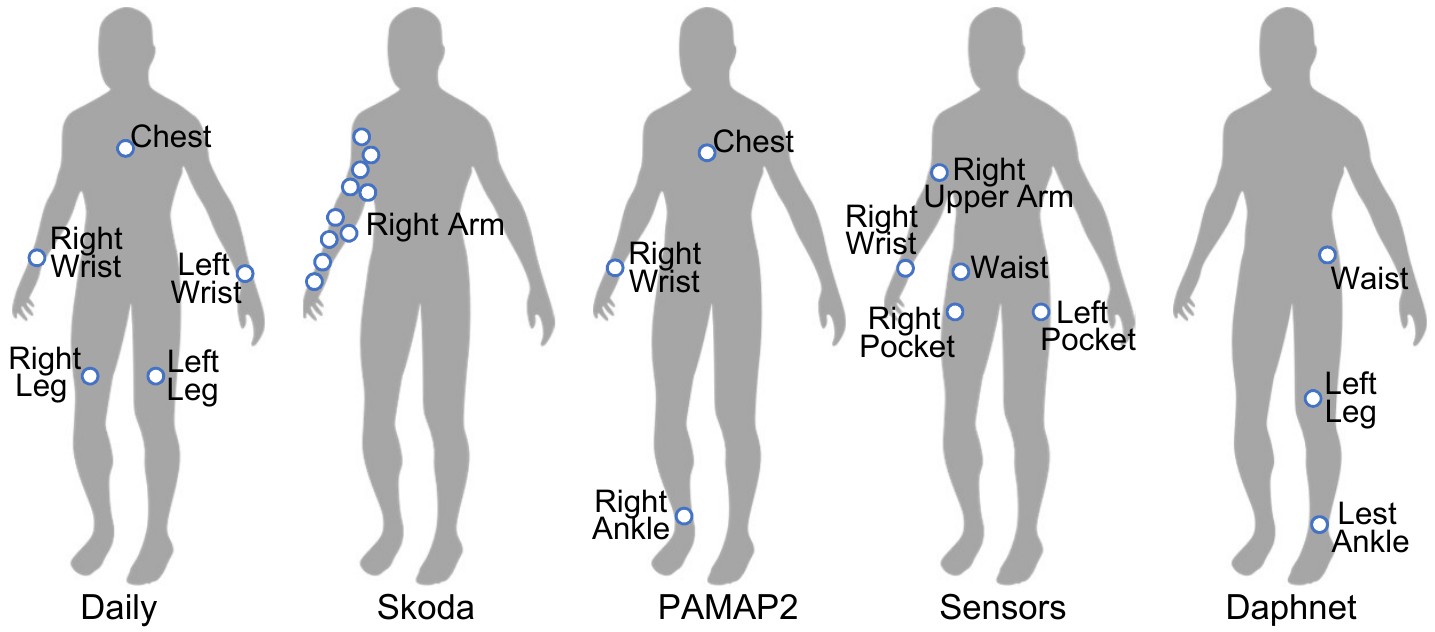}
	\caption{Worn locations of the five datasets (Daily~\cite{barshan2014recognizing}, Skoda~\cite{zappi2012network}, PAMAP2~\cite{reiss2012introducing}, Sensors~\cite{shoaib2014fusion}, and Daphnet~\cite{bachlin2010wearable}).}
	\label{fig:sensor_loc}       
\end{figure}

As summarized in Table~\ref{tab:dataset}, we selected five publicly available datasets for the method validation. These datasets are collected in various contexts by different research groups, including different sensor positions on the human body, different sampling rates, and different numbers of subjects. In addition, the five datasets include activities with different levels of classification difficulties, for example, the relatively more discriminative activities~\cite{shoaib2014fusion} such as walking, sitting, 
and complex activities in special scenarios such as the manipulative gestures performed in a car maintenance workshop~\cite{zappi2012network}. Figure~\ref{fig:sensor_loc} shows the senor locations on a human body for the five datasets. By leveraging these five different datasets, we are able to test the effectiveness and robustness of our approach.

\begin{table}[h]\footnotesize
	\centering
	\caption{Information of the five public datasets.}
	\label{tab:dataset}
	\begin{threeparttable}
		\begin{tabular}{cp{1.0cm}p{1.5cm}p{1.5cm}p{1.0cm}p{1.5cm}p{1.5cm}}
			\hline\noalign{\smallskip}
			Datasets & \#Sensors & Modalities & Number of Channels & Rate (Hz) & Number of Activities & Number of Subjects\\
			\noalign{\smallskip}\hline\noalign{\smallskip}
			Daily~\cite{barshan2014recognizing} & 5 & $A,G,M$& 9 & 25 & 19 & 8\\
			Skoda~\cite{zappi2012network} & 10 & $A$& 3 & 98 & 10 & 1\\
			PAMAP2~\cite{reiss2012introducing} & 3& $A,G,M$ & 9 & 100 & 12 & 9\\
			Sensors~\cite{shoaib2014fusion} & 5& $A,\bar{A},G,M$ & 12 & 50 & 7 & 10\\
			Daphnet~\cite{bachlin2010wearable} & 3& $A$ & 3 & 64 & 2 & 10\\
			\noalign{\smallskip}\hline
		\end{tabular}
		\begin{tablenotes}
			\item Note: $A,\bar{A},G,M$ represent the modalities of acceleration, gravity-removed acceleration, angular velocity, and magnetic field, respectively.
		\end{tablenotes}
		
	\end{threeparttable}
\end{table}

\textbf{Daily and Sports Activity Dataset}~\cite{barshan2014recognizing} This dataset is composed by IMU data of 19 daily and sports activities ((1) sitting, (2) standing, (3-4) lying on the back and on the right side, (5-6) ascending and descending stairs, (7) standing in an elevator still, (8) moving around in an elevator, (9) walking in a parking lot, (10-11) walking on a treadmill with a speed of 4 km/h (in flat and 15 deg inclined positions), (12) running on a treadmill with a speed of 8 km/h, (13) exercising on a stepper, (14) exercising on a cross trainer, (15-16) cycling on an exercise bike in horizontal and vertical positions, (17) rowing, (18) jumping, (19) playing basketball.), 
captured by five IMU devices (worn on the torso, right arm, left arm, right leg, and left leg, respectively), and the activities are performed by 8 different subjects.

\textbf{Skoda Dataset}~\cite{zappi2012network} This dataset contains 10 manipulative activities performed in a car maintenance scenario by a single subject (e.g., the user blocks an opened hood with a stick, and the user grabs the steering wheel and turns it).
The dataset has signal recordings from both the left and right arms but they are not synchronized for validation. Therefore, in this study, we focus on signals from 10 sensors worn on the subject's right arm. 

\textbf{PAMAP2 Dataset}~\cite{reiss2012introducing} This dataset has 12 human activities ((1) lying, (2) sitting, (3) standing, (4) walking, (5) running, (6) cycling, (7)Nordic walking, (8) ascending stairs, (9) descending stairs, (10) vacuum cleaning, (11) ironing and rope jumping) 
captured by three IMU sensors (worn on the wrist, chest and ankle, respectively), and
the activities are performed by 9 different subjects. 

\textbf{Sensors Activity Dataset}~\cite{shoaib2014fusion} This dataset includes 7 human activities ((1) biking, (2) downstairs, (3) jogging, (4) sitting, (5) standing, (6) upstairs, and (7) walking) 
captured by five IMU sensors (one in the the right jeans pocket, one in the left jeans pocket, one on the belt position towards the right leg using a belt clip, one on the right upper arm, one on the right wrist), and
the activities are performed by 10 different subjects.

\textbf{Daphnet Freezing of Gait Dataset}~\cite{bachlin2010wearable} This dataset contains 3 wearable wireless acceleration sensors at the hip and leg of Parkinson's disease patients that experience freeze of gait (FoG) during walk tasks. This dataset has two classes, FoG and \lq no freeze\rq, 
captured by three sensors (worn at the ankle (shank), on the thigh just above the knee, and on the hip, respectively), and 
the activities are collected from 10 different patients.

\subsection{Evaluation Metrics}
Regarding to evaluation metric, the leave-one-out evaluation policy is conducted.
In the leave-one-out evaluation, the samples from $N_{subject}-1$ out of $N_{subject}$ subjects are used for training, and the samples of the left one subject are reserved for testing.
We employ several commonly used metrics~\cite{Goodfellow-et-al-2016} to evaluate the classification performance, which are listed as follows:
\begin{itemize}
\item Accuracy
\begin{equation}
Accuracy=\frac{\sum^N_n1(\hat{y}_n=y_n)}{N}
\end{equation}
\item Precision and Recall
\begin{equation}
\begin{split}
	Precision=\frac{TP}{TP+FP} \\
	Recall=\frac{TP}{TP+FN}
\end{split}
\end{equation}
\item $F_1$ score
\begin{equation}
F_1=2\cdot\frac{Precision\cdot Recall}{Precision+Recall}
\end{equation}
\end{itemize}
where $1(\cdot)$ is an indicator function. For a certain class $y_i$, True Positive (TP) is defined as a sample of class $y_i$ that is correctly classified as $y_i$; 
False Positive (FP) means a sample from a class other than $y_i$ is misclassified as $y_i$; False Negative (FN) means a sample from the class $y_i$ is misclassified as another \lq not $y_i$\rq\ class. $F_1$ score is the harmonic mean of Precision and Recall, which ranges in the interval [0,1].

\subsection{Implementation Details}
The DNN architectures described in Section~\ref{method} are constructed using TensorFlow~\cite{tensorflow2015-whitepaper} and Keras libraries.
The SGD optimizer is used in training, with the momentum of 0.9, the learning rate of 0.001 and the regularizer coefficient of 1e-5. 
We use a workstation with one 12-core Intel Xeon processor, 64GB of RAM and two Nvidia Geforce 1080 Ti graphic cards for the training jobs. 

\subsection{Evaluation of Different Signal Representation Methods}

To evaluate how the design of signal representation affects the model performance,
comparisons have been made among methods using images of (1) raw signals ($I^{RS}$), (2) Discrete Cosine Transform ($I^{DCT}$),
and (3) Discrete Fourier Transform ($I^{DFT}$).
Table~\ref{tab:compare_transform_method} shows the performance of activity recognition with various designs of input images.

\begin{table}[h]\footnotesize
\begin{center}
\begin{threeparttable}
\caption{Performance (\%) comparison of different signal representation methods on the Daily dataset.}

\begin{tabular}{cccccc}
\hline
Methods & Input Size & Accuracy & Precision & Recall & F Score \\
\hline
$I^{RS}$ & $C\times T$& 67.57 & 64.50 & 67.57 & 61.78 \\

$I^{RS}$ $
\underrightarrow{(DCT)}$ $I^{DCT}$ & $C\times T$& 90.36 & 91.85 & 90.36 & 89.44 \\
$I^{RS}$ $
\underrightarrow{(DFT)}$ $I^{DFT}$ & $C\times (T/2)$& 90.37 & 91.86 & 90.37& 89.82   \\
\hline
\end{tabular}
\begin{tablenotes}
\item Note: $I^{RS}$, $I^{DCT}$ and $I^{DFT}$ represent image representations of raw signals, DCT and DFT, respectively. $C$ and $L$ denote the number of signal channels and the number of time frames in a signal segment, respectively.
\end{tablenotes}

\label{tab:compare_transform_method}
\end{threeparttable}

\end{center}
\end{table}

The proposed signal representation method $I^{DFT}$ achieves the highest recognition performance. The performance decreases when we use the image of raw signals $I^{RS}$ directly or replace the Discrete Fourier Transform with the Discrete Cosine Transform ($I^{DCT}$).
Therefore, $I^{DFT}$ is selected for the signal representation.
Another reason for choosing DFT over DCT is that DFT is symmetric, and only half the image size after remove its symmetric part,  which will reduce the complexity of the DNN model and has a better computational efficiency. It saves 50\% of the first-layer computation over a DCT.

\subsection{Evaluation of the Length of the Signal Segment}
When sampling the signals (the sampling procedure is discussed in Section~\ref{sec:signal_preprocess}), as shown in Figure~\ref{fig:sample}, there are two parameters to choose, the length of the segment ($T$) and the stride ($\Delta_t$), which determine how much information the model can digest at each time, and how much shared overlap between two segments, respectively. Here the question is what should be the optimal length and stride for sampling to identify an activity. Table~\ref{tab:compare_sample_length} presents the performance comparison of different settings of length and stride evaluated on the validation dataset.

\begin{table}[h]\footnotesize
\begin{center}
\begin{threeparttable}
\caption{Performance (\%) comparison of different settings of segment length and stride on the Daily dataset.}
\begin{tabular}{cccccc}
\hline
Length & stride & Accuracy & Precision & Recall & F Score \\
\hline
32 & 8  & 92.39 & 93.62 & 92.39 & 91.55 \\
32 & 16  & 92.37 & 93.74 & 92.37 & 91.91 \\
32 & 24 & 90.07 & 91.31 & 90.07 & 89.06 \\
64 & 16 & 90.37 & 91.86 & 90.37& 89.82\\
64 & 32 & 86.63 & 88.47 & 86.63 & 85.24 \\ 
96 & 24 & 89.11 & 90.87 & 89.11 & 88.23 \\
125 & --* & 85.43 & 87.83 & 85.43 & 84.11 \\
\hline
\end{tabular}
\begin{tablenotes}
\item *Since the sequence length of the Daily dataset is $125$, the stride value is absent in the last row.
\end{tablenotes}
\label{tab:compare_sample_length}
\end{threeparttable}
\end{center}
\end{table}

The accuracy decreases when increasing the segment length, because longer length could have multiple repeated patterns in each segment, which makes it harder for the DNN model to learn the most discriminative features. Also, longer segment length leads to less segments, i.e., less training data, which affects the training effect.
In terms of stride, short strides can have better performance. This is because the model tends to look into the data more precisely with a shorter stride. 
Therefore, we choose the parameter setting, $T=32$ and $\Delta_t=8$, for the following experiments.

\subsection{Evaluation of the Effectiveness of the Fusion Mechanism}
In terms of data fusion, as shown in Figure~\ref{fig:overview}, the information flows are fused at two places: fusion of multi-channel data of a specific sensor in the sensor-wise feature extraction module (Sections~\ref{sec:sensor-wise_module}) and fusion of multi-sensor data in the inter-sensor feature extraction module (Section \ref{sec:inter-sensor}). The fusion mechanism is realized using convolutional operations with different receptive fields, i.e., 2D kernels of different sizes. When a 2D kernel moves over an area, the hovered information is fused with the summation of point-wise multiplications.
Here to validate the effectiveness of the fusion mechanism, we compare it with a method using 1D convolutions which does not include fusion functionalities. 
The results are listed in Table~\ref{tab:compare_correlation_module}.
We can see that, the performance drops dramatically after ignoring the fusion, which demonstrates the the designed fusion mechanism plays a vital role in identifying an activity.
\begin{table}[h]\footnotesize

\begin{center}
\begin{threeparttable}
\caption{Performance (\%) evaluation of the effectiveness of the fusion mechanism.}
\begin{tabular}{lcccc}
\hline
Method & Accuracy & Precision & Recall & F Score \\
\hline
Without Fusion Mechanism* & 62.95 & 63.99 & 62.95 & 58.73 \\
With Fusion Mechanism & 92.37 & 93.74 & 92.37 & 91.91 \\
\hline
\end{tabular}
\begin{tablenotes}
\item * 1D convolutions along each row of the feature maps to ignore the fusion mechanism.
\end{tablenotes}
\label{tab:compare_correlation_module}
\end{threeparttable}
\end{center}
\end{table}

\subsection{Evaluation of Different Fusion Methods}
\begin{figure}[H]
\centering
\includegraphics[width=1.\textwidth]{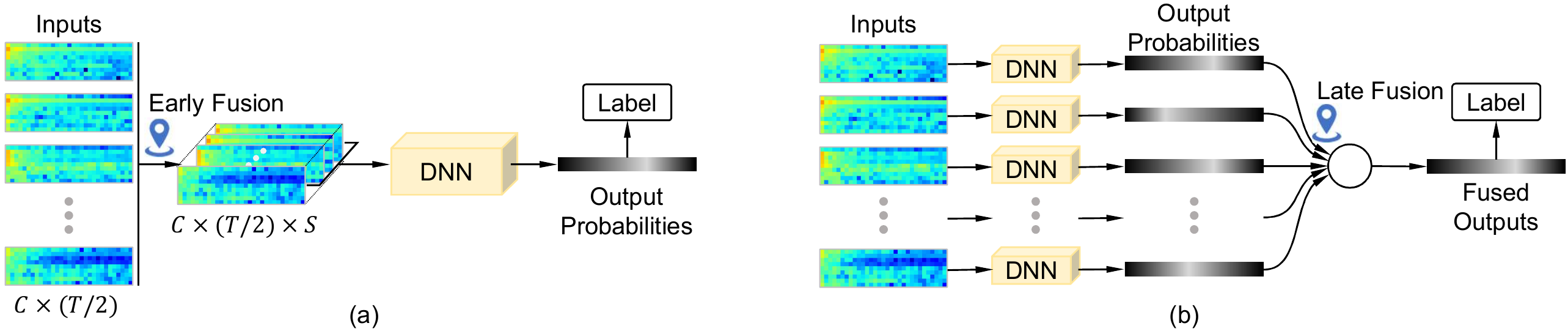}
\caption{Architectures of different fusion methods: (a) early fusion and (b) late fusion.}
\label{fig:fusion_method}       
\end{figure}

In this experiment, we compare our \textbf{attention-based fusion} method with two other fusion methods (early fusion and late fusion), whose architectures are presented in Figure~\ref{fig:fusion_method}.

\textbf{Early fusion} fuses information in the input phase. As shown in Figure~\ref{fig:fusion_method}(a), all the $S$ inputs are stacked to generate a single input with the size of $C\times (T/2)\times S$. Then, the integrated input is fed into a DNN model.

\textbf{Late fusion} fuses information in the inference phase. As shown in Figure~\ref{fig:fusion_method}(b), all the $S$ sensor inputs are learned by different DNN models individually. Then, their inferred output probabilities are fused to generate a final output.

The performance comparison of different fusion methods is listed in Table~\ref{tab:compare_fusion}.
For early fusion, the inputs are integrated before feature extraction modules of the DNN model, which lacks individual understanding of signal from each sensor. Later fusion relies on individual sensor to learn the features and achieves higher performance, but it doesn't have the ability to look into the deep correlations among different sensors as attention fusion does. Overall, the attention fusion achieves the best results.

\begin{table}[h]\footnotesize
\begin{center}
\caption{Performance (\%) comparison of different fusion methods.}
\begin{tabular}{ccccc}
\hline
Method & Accuracy & Precision & Recall & F Score \\
\hline
\hline
Early Fusion & 89.62 & 90.63 & 89.62 & 88.86 \\
\hline
Late Fusion & 91.57 & 92.30 & 91.57 & 90.43 \\
\hline
Attention Fusion & 92.37 & 93.74 & 92.37 & 91.91 \\
\hline
\end{tabular}
\label{tab:compare_fusion}
\end{center}
\end{table}

\subsection{Comparison with the State-of-the-Art Methods}
In this subsection, we compare our results with the state-of-the-art performance on the five public datasets. 
The comparison is summarized in Table~\ref{tab:comparison}. 
We also evaluate our model without the attention mechanism, in which the sensor attention module is removed.
Overall, our proposed model achieves higher accuracy than the other methods, 
which is attributed to two factors: a more effective signal representation method exposing the hidden patterns and an attention-based sensor fusion model extracting the most discriminative features. 

\begin{table}[h]\footnotesize
\begin{center}
\caption{Performance (\%) comparison of existing models on the five public datasets. \lq --\rq\ denotes that the value is not reported in the paper.}
\begin{tabular}{lccccc}
\hline
Approach & Daily & Skoda & PAMAP2 & Sensors & Daphnet \\
\hline
Zhang et al. (2015)~\cite{zhang2015recognizing} &90.60&--& --&--&--\\
Hammerla et al. (2016)~\cite{hammerla2016deep} &--&--& 93.70&--&76.00\\
Ord{\'o}{\~n}ez et al. (2016)~\cite{ordonez2016deep} &--& 95.80 & -- &--&--\\
Guan et al. (2017)~\cite{guan2017ensembles} &--& 92.40 & 85.40 &--&--\\
Xi et al. (2018)~\cite{xi2018deep} &--&--& 93.50&--&-- \\
Murahari and PI{\"o}tz (2018)~\cite{murahari2018attention} & -- & 91.30 & 87.50 & -- & --\\
Zeng et al. (2018)~\cite{Zeng:2018:UIR:3267242.3267286} & -- &93.81 & 89.96 & -- & 83.73\\
Cao et al. (2018)~\cite{cao2018optimizing} & 78.48 &-- & -- & -- & --\\
Moya Rueda et al. (2018)~\cite{moya2018convolutional} & -- &-- & -- & 93.68 & --\\
Mohammad et al. (2018)~\cite{mohammad2018deep} & -- &91.20 & -- & -- & --\\
Shakya et al. (2018)~\cite{shakya2018comparative} & -- &-- & -- & 99.16 & --\\
Xu et al. (2019)~\cite{xu2019innohar} &--&--& 93.50&--&-- \\
\hline
Our model without attention & {88.55} & 94.16 &93.14 & 97.36 &  89.81\\
\textbf{Our model with attention} & \textbf{92.37} & \textbf{95.84} & \textbf{94.85} & \textbf{99.27} & \textbf{91.02}\\
\hline
\end{tabular}
\label{tab:comparison}
\end{center}
\end{table}

\begin{figure}[H]
\centering
\includegraphics[width=.9\textwidth]{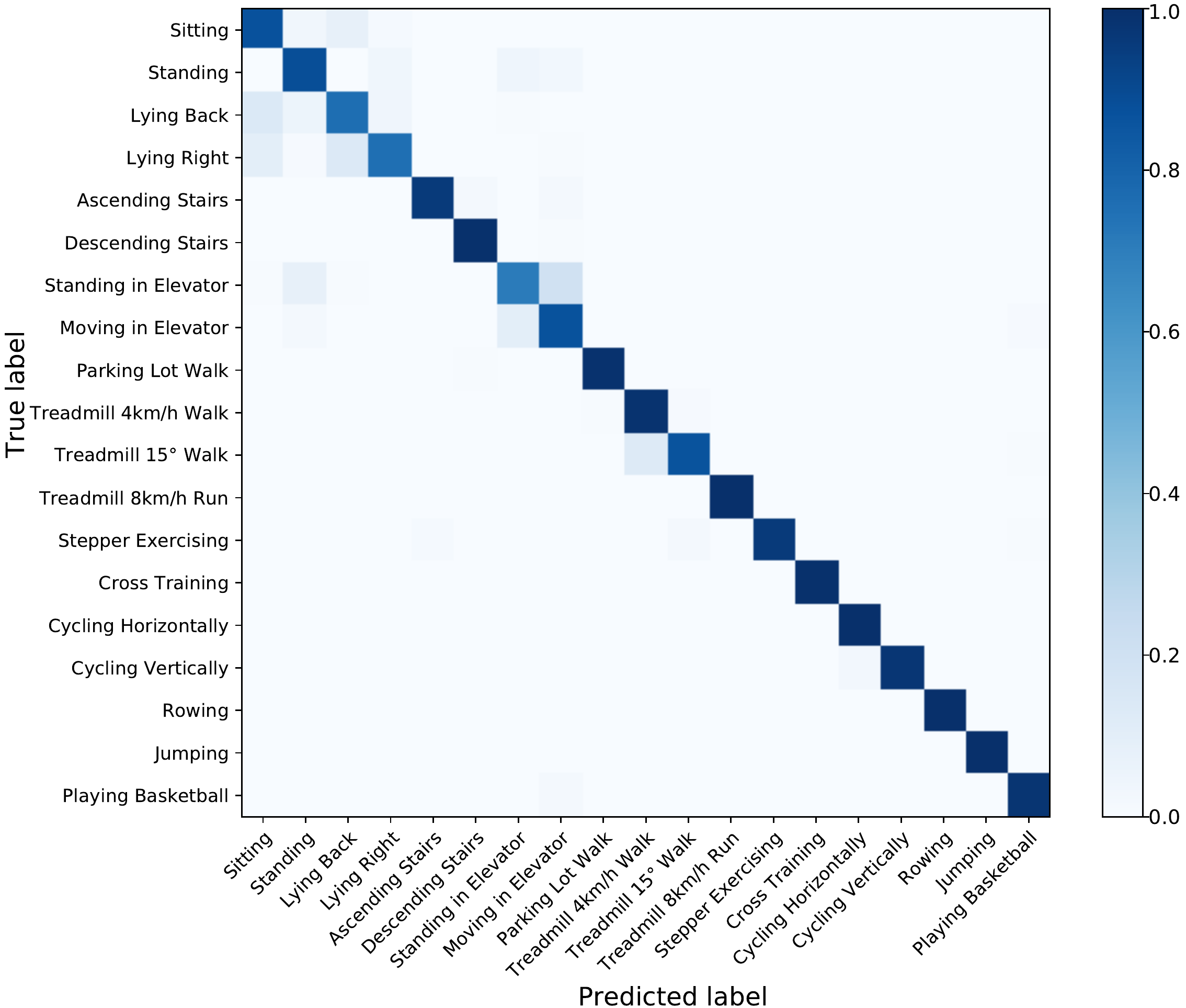}
\caption{Normalized confusion matrix of the Daily dataset.}
\label{fig:confusion_matrix}       
\end{figure}

Figure~\ref{fig:confusion_matrix} shows the normalized confusion matrix of the Daily dataset.
We can see that most of the activities are successfully classified.
Failures occur in classifying the confusing groups: e.g., (1) sitting, lying on the back, and lying on the right side; (2) standing, standing in the elevator, and moving in the elevator; (3) treadmill walking in flat position and treadmill walking in 15 deg inclined position.
By reviewing the failure cases, we find that the high similarity within the confusing groups makes it difficult to distinguish them from others, and the significant subject-wise difference for the same activity makes it difficult to learn this kind of unseen variations beforehand.

\subsection{Visualization of the Learned Sensor Attention}
\begin{figure}[H]
\centering
\includegraphics[width=1.\textwidth]{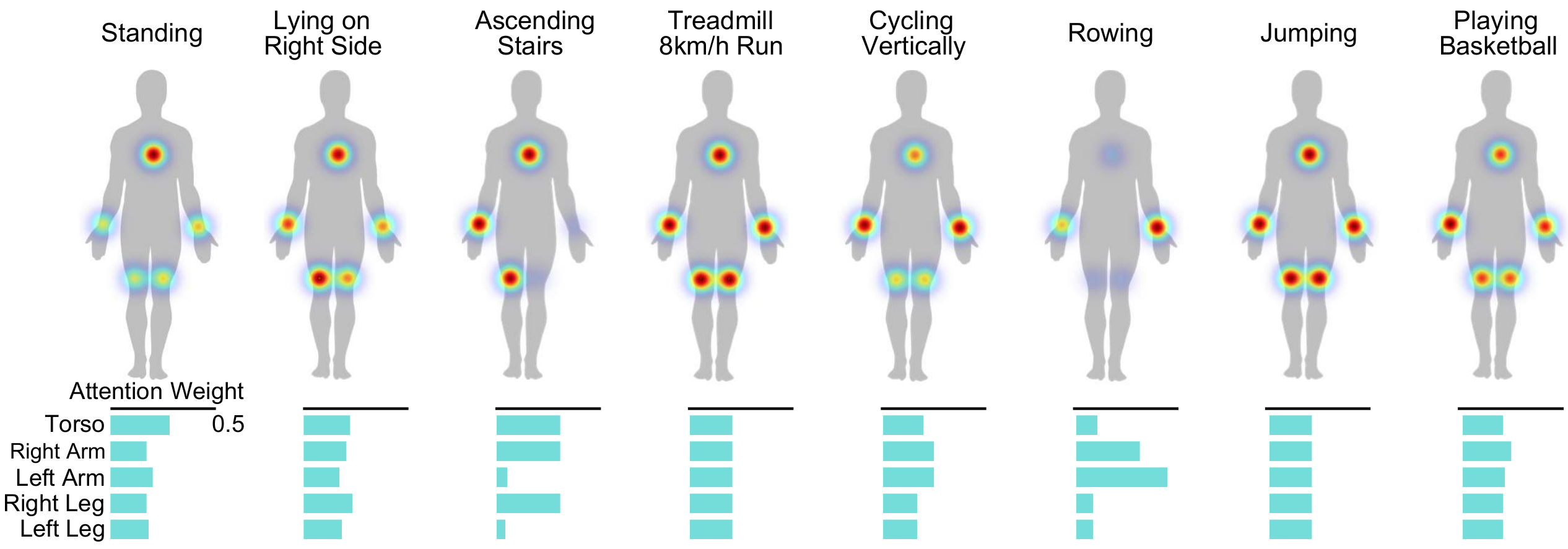}
\caption{Examples of the importances of sensor at different body locations. The heatmaps represent the importance and the attention weights of all sensors are illustrated in the lower barchart.}
\label{fig:vis}       
\end{figure}

In this section, we analyze and visualize the learned attention, i.e., attention weights, of sensors at different body locations. 
The attention vector $\hat{\vec{a}}_{softmax}$ (Eq.~\ref{eq:attention}) is extracted from a well-trained model and each element of this vector is represented as a heatmap.
A few examples of the sensor attention trained on the Daily dataset are shown in 
Figure~\ref{fig:vis}, where \lq hotter\rq\ colors represent larger values while \lq colder\rq\ colors represent smaller ones on the blue-red heatmaps.
We can see that different activities shows different attention distributions. For example, the \lq rowing\rq\ activity has larger attention weights for sensors worn on the arms, because the motion intensities of the arms are larger than other body parts. While for activities such as \lq running\rq, \lq jumping\rq, and \lq playing basketball\rq, the attention is more evenly distributed across different sensors, because these activities involve the whole body.
This visualization shows that our model is able to focus on the critical body parts based on their importance when identifying activities. 

\subsection{Visualizing the Class Activation Map}

\begin{figure*}[h]
\begin{center}
\includegraphics[width=1.\linewidth]{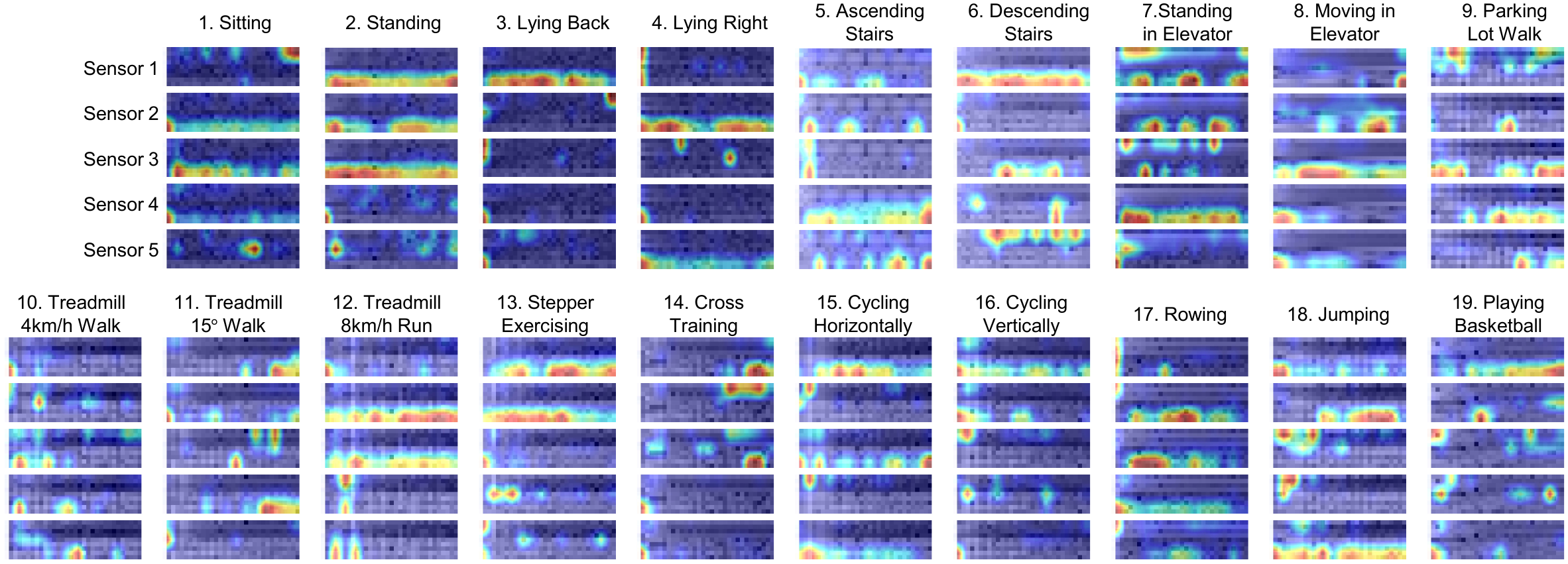}
\end{center}
\caption{Examples of Class Activation Map (CAM) Visualization. (Best in color)}
\label{fig:cam}
\end{figure*}

To have a more intuitive understanding of which regions of an input image are more discriminative to activate our model to its final inference, we visualize the class activation map (CAM), 
which is a 2D grid of scores associated with a specific output class, computed for every region in an input image, indicating the importance of each region in regard to the class under consideration. 
A set of CAM examples are shown in Figure~\ref{fig:cam}, where the generated heatmaps are overlaid onto the input images. We can see that the model automatically learns the most discriminative regions in an input image and different activities use different regions (i.e., different signal channels and frequency characteristics) in identifying their categories.

\section{Conclusions and Remarks}
\label{conclusion}
In this paper, we propose a novel approach of attention-based sensor fusion for Human Activity Recognition (HAR) using Inertial Measurement Unit (IMU) signals obtained from multiple sensors worn at different body locations.
For signal representation, a simple yet effective pipeline for feature transform is designed to represent the input signals of each sensor as images in the frequency domain.
Having the formatted images as inputs, a sensor-wise feature extraction module is developed to extract the most discriminative features of signals from individual sensors with Convolutional Neural Networks (CNNs), and to generate a vector representation for each sensor.
Then, a sensor attention mechanism is developed to learn the importance of sensors at different body locations and to create an attentive feature representation.
After that, an inter-sensor feature extraction module is applied to learn the inter-sensor correlations, which are connected to a classifier to output the predicted classes of activities. 
This attention-based model is able to learn the importance of sensors at different body locations, yielding a more comprehensive understanding of the human activity.
The proposed approach is evaluated on five publicly available datasets and it demonstrates superior performance than the state-of-the-art methods.

To further improve the current approach for higher performance and practical applications, some directions for future study can be considered, such as exploring data augmentation techniques to introduce more variations to the collected data, 
experimenting other methods of signal preprocessing and representation to fully exploit the discriminative information within the recorded signals,
and developing channel-wise attention mechanism to look into the importance of each individual channel for a sensor at a specific location.
In addition, cross-dataset recognition approach can be explored.

\section*{Acknowledgement}
This research work is supported by the National Science Foundation under Grant Cyber-Physical Sensing (CPS) Synergy project CMMI-1646162 and National Robotics Initiative (NRI) project CMMI-1954548, and also by the Intelligent Systems Center at Missouri University of Science and Technology. Any opinions, findings, and conclusions or recommendations expressed in this material are those of the authors and do not necessarily reflect the views of the National Science Foundation.

\bibliography{refs}

\end{document}